\documentclass[11pt]{article}

\usepackage[preprint]{acl}

\usepackage{times}
\usepackage{latexsym}
\usepackage{url}
\usepackage{amsmath} 

\usepackage{wrapfig}
\usepackage{hyperref}
\usepackage{url}
\usepackage{tikz}
\usepackage{multirow}
\usepackage{booktabs}
\usepackage{enumitem}
\usepackage{xspace}
\usepackage{adjustbox}
\usepackage{tabularray}
\usepackage{booktabs,tabularx}
\usepackage{wrapfig}
\usepackage{verbatim}
\usepackage{fancyvrb}
\usepackage{fvextra}
\usepackage{scalerel}
\usepackage{pifont}
\usepackage{subcaption}
\usepackage{minted} 
\usepackage{amsmath, amssymb} 
\usepackage{adjustbox}        
\usepackage{listings}

\usepackage[most]{tcolorbox}

\usepackage{xcolor}

\definecolor{keywordblue}{RGB}{0, 0, 180}
\definecolor{keywordred}{RGB}{180, 0, 0}

\usepackage[T1]{fontenc}

\usepackage[utf8]{inputenc}

\usepackage{microtype}

\usepackage{inconsolata}

\usepackage{graphicx}
\usepackage{booktabs}
\usepackage{multirow}
\usepackage{amsmath}
\usepackage{amssymb}
\newcommand{\datasetname}{{\textbf{PhysLeanData}}}
\newcommand{\ourmodel}{{\textbf{PhysProver}}}
\newcommand{\ourframework}{{\textbf{Physics Prover Framework}}}

\title{PhysProver: Advancing Automatic Theorem Proving for Physics}

\author{
 \textbf{Hanning Zhang\textsuperscript{*1}},
 \textbf{Ruida Wang\textsuperscript{*1}},
 \textbf{Rui Pan\textsuperscript{*1}},
 \textbf{Wenyuan Wang\textsuperscript{*2}},
\\
 \textbf{Bingxu Meng\textsuperscript{1}},
 \textbf{Tong Zhang\textsuperscript{1}}
\\
 \textsuperscript{1}University of Illinois Urbana-Champaign
 \\
 \textsuperscript{2}Rutgers University \\
 \texttt{\{hanning5, ruidaw, ruip4, bingxum2, tozhang\}@illinois.edu}
 \\
 \texttt{ww462@scarletmail.rutgers.edu}
}

\begin{document}
\maketitle

\begin{abstract}

The combination of verifiable languages and LLMs has significantly influenced both the mathematical and computer science communities because it provides a rigorous foundation for theorem proving. Recent advancements in the field provide foundation models and sophisticated agentic systems pushing the boundaries of formal mathematical reasoning to approach the natural language capability of LLMs~\cite{chen2025seed15}. However, little attention has been given to the formal physics reasoning, which also heavily relies on similar problem-solving and theorem-proving frameworks. To solve this problem, this paper presents, to the best of our knowledge, the first approach to enhance formal theorem proving in the physics domain. We compose a dedicated dataset \datasetname{} for the task. It is composed of theorems sampled from PhysLean~\cite{tooby2025heplean} and data generated by a conjecture-based formal data generation pipeline. In the training pipeline, we leverage DeepSeek-Prover-V2-7B, a strong open-source mathematical theorem prover, and apply Reinforcement Learning with Verifiable Rewards (RLVR) to train our model \ourmodel{}. Comprehensive experiments demonstrate that, using only $\sim$5K training samples, \ourmodel{} achieves an overall \textbf{2.4\%} improvement in multiple sub-domains. Furthermore, after formal physics training, we observe \textbf{1.3\%} gains on the MiniF2F-Test benchmark, which indicates non-trivial generalization beyond physics domains and enhancement for formal math capability as well. The results highlight the effectiveness and efficiency of our approach, which provides a paradigm for extending formal provers outside mathematical domains. To foster further research, we will release both our dataset and model to the community.

\end{abstract}
\section{Introduction}
\label{sec:introduction}

\begin{figure*}[t]
    \vspace{-0.0in}
    \centering
    \includegraphics[width=0.99\linewidth]{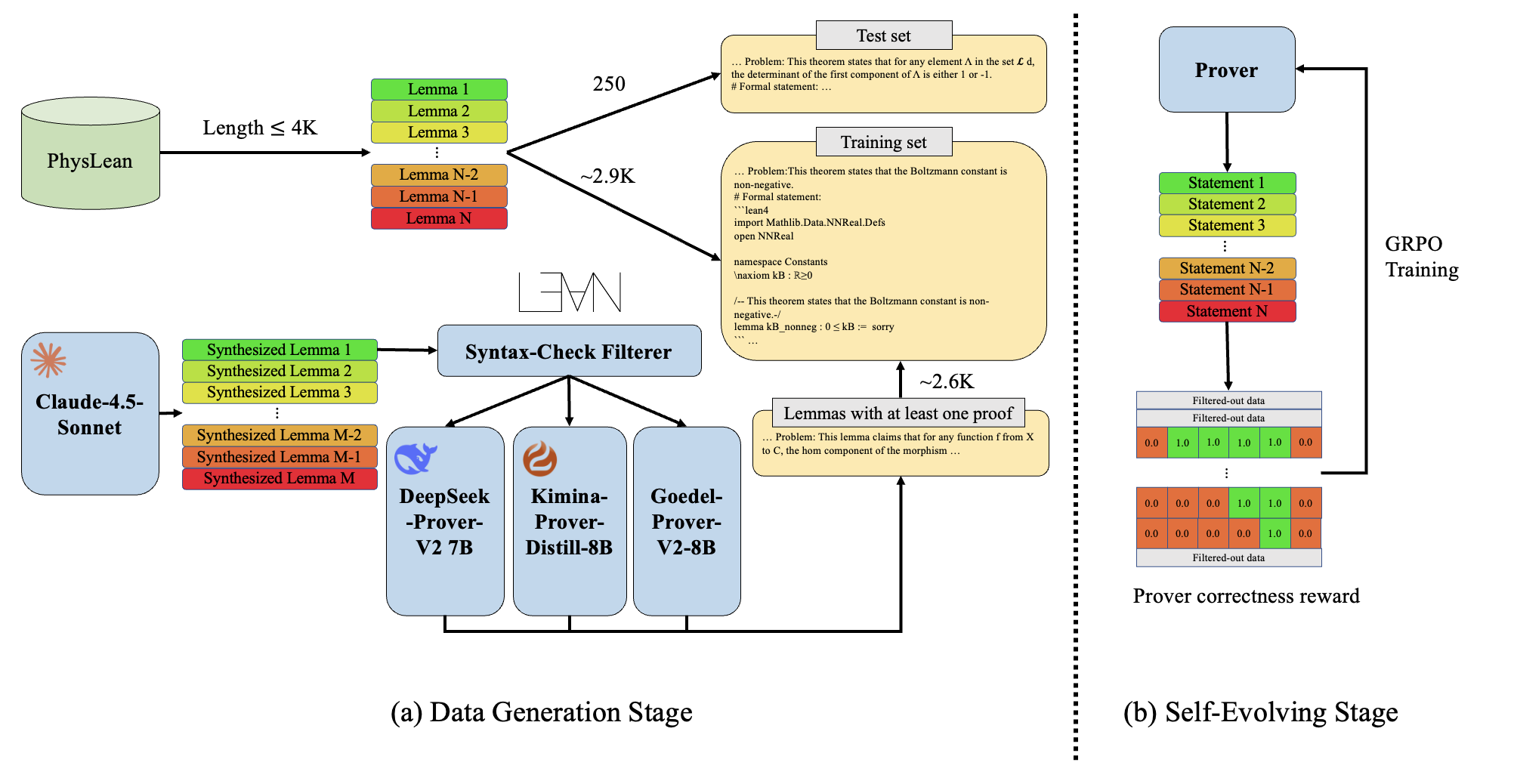}
    \caption{
    \textbf{\ourframework}: (a) Data Generation Stage: the training set comprises 5,541 physics statements from both PhysLean~\citep{tooby2025heplean} and synthetic lemmas from Claude-4.5-Sonnet, where the latter are further filtered by Lean syntax and proof existence checks. (b) Self-Evolving Stage: after obtaining the training set, GRPO~\citep{shao2024deepseekmathgrpo} is adopted to train the base prover models, with reward signals of proof correctness provided by Lean.}
    \label{fig:main}
    \vspace{-0.2in}
\end{figure*}

Formal reasoning has long been recognized as a cornerstone of human intelligence and a critical domain in machine learning research~\cite{P-868}. With the recent advancements in Large Language Models (LLMs), much research has investigated their applications in formal theorem proving. They explored domains from training foundation models~\cite{lin2025goedelproverv2, ren2025deepseekproverv2, wang2025ma} and specialized agent framework~\cite{wang2025gar, chen2025seed15, varambally2025hilbertrecursivelybuildingformal}. Among these, math theorem proving in Lean4~\cite{10.1007/978-3-030-79876-5_37} has emerged as one of the most extensively studied areas~\cite{wang2024theoremllamatransforminggeneralpurposellms, lin2025goedelproverfrontiermodelopensource, xin2024deepseekproveradvancingtheoremproving}. Researchers typically start from a general-purpose LLM, employing Supervised Fine-Tuning (SFT) and Reinforcement Learning (RL) to enhance the formal reasoning capability. This approach has achieved strong results on formal math benchmarks, such as MiniF2F~\cite{zheng2022minif2fcrosssystembenchmarkformal} and PutnamBench~\cite{tsoukalas2024putnambenchevaluatingneuraltheoremprovers}\footnote{\url{https://github.com/hanningzhang/PhysProver}}. 

Previous works have demonstrated that developing expert models for Lean4 theorem proving demands substantial training data and a large amount of GPU hours. For instance, DeepSeek-Prover~\cite{xin2024deepseekproveradvancingtheoremproving} applies a 120B math-related tokens continue pretraining and 8M formal statements with proofs to train an expert prover. Similarly, Goedel-Prover~\cite{lin2025goedelproverfrontiermodelopensource} applies expert iteration on more than 1 million formal statements. Despite these advancements, formal theorem proving faces significant challenges due to a scarcity of high-quality data that is able to give the model a general formal reasoning capability, rather than focusing on a narrow field~\cite{li2025lean4physics}.

While significant progress has been made in mathematical theorem proving, the formal physics domain remains largely overlooked. Physics, with its reliance on rigorous mathematical foundations and formal derivations, offers a natural yet under-explored extension to formal reasoning. \citet{li2025lean4physics} highlights that SOTA theorem-proving models perform poorly in physics-related tasks but fail to propose methods for improvement. 

To settle this gap, we, as far as we are concerned, take the first step toward enhancing theorem proving in the physics domain by constructing a specialized data pipeline and employing Reinforcement Learning with Verifiable Rewards (RLVR). The overview of our framework can be found in Figure~\ref{fig:main}. Specifically, we collect foundational theorems and lemmas from the open-source repository of PhysLean~\cite{tooby2025heplean}, which contains Lean4-based results across advanced physics domains such as Quantum Field Theory and String Theory. The extracted data, along with their headers, are divided into the training and testing sets. To augment the training dataset, we apply Claude-4.5 to generate additional conjectures based on the dataset. Subsequently, we apply formal LLMs to annotate these conjectures, thereby formulating the Basic Physics Lean training dataset, which contains approximately 5K training samples and 250 testing samples.

With the dataset, we leverage RLVR~\cite{lambert2025tulu3pushingfrontiers} to enhance physical theorem-proving capability using the GRPO algorithm. Our evaluation demonstrates consistent improvement across multiple physics domains and achieves a \textbf{2.4\%} of overall improvements compared to SOTA math provers on the testing dataset. Furthermore, when tested on the Out-of-Distribution (OOD) MiniF2F benchmark~\cite{zheng2022minif2fcrosssystembenchmarkformal}, \ourmodel{} achieved over 1\% of improvement compared to the base model under pass@16. It demonstrates the effectiveness of our approach and shows that physics dataset training can enhance the formal math capability of the model.

We summarize our contribution as follows:
\begin{enumerate}
    \item Introducing the first methods specifically designed to train the formal theorem provers for physics.
    \item Developing and releasing a compact and comprehensive small dataset, along with a conjecture synthesis pipeline for physical theorems to benefit the research community.
    \item Training a formal physics prover that outperforms the SOTA model and achieves superior performance in both physics and mathematical theorem proving.
\end{enumerate}

\section{Related Works}\label{sec:related_works}

\subsection{Formal Math Reasoning}\label{related_works:math}

Formal math reasoning involves representing mathematical components in a computer-verifiable format. It reduces the ambiguity and establishes a rigorous foundation for logical reasoning. Over the past decades, researchers have developed numerous Formal Languages (FLs) based on two primary theoretical frameworks. The first relies on dependent type languages, such as Lean~\cite{de2015lean, moura2021lean} and Coq~\cite{coq1996coq}, where formal verification is achieved through a small kernel to perform type checking. The second line utilizes higher-order logic to quantify functions and predicates. This line of work is represented by languages such as Isabelle~\cite{paulson1994isabelle}, HOL, and HOL Light~\cite{harrison2009hol}. Among the above languages, Lean4~\cite{moura2021lean} has gained significant attention due to its expressiveness and extensive Mathlib4 repository, which encompasses almost all major mathematical domains.

The rise of LLMs has accelerated the advancements in formal proving tasks. Researchers have compiled extensive datasets of mathematical theorems and proofs\cite{wang2025ma, lin2025goedelproverfrontiermodelopensource, dong2025stpselfplayllmtheorem}, which provide a robust foundation for model training. Building on these resources, increasingly sophisticated models have emerged. Early efforts, such as Expert Iteration~\cite{polu2022formal}, employed iterative annotation using LLMs to enhance the training data. Open-source frameworks like DeepSeek-Prover~\cite{xin2024deepseekproveradvancingtheoremproving} and TheoremLlama~\cite{wang2024theoremllamatransforminggeneralpurposellms} further advanced the formal provers. More recently, RLVR has enabled Long CoT training for formal theorem proving, which works like MA-LoT~\cite{wang2025ma}, Kimina-Prover~\cite{wang2025kiminaprover}, DeepSeek-Prover-V2~\cite{ren2025deepseekproverv2}, and Goedel-Prover-V2~\cite{lin2025goedelproverv2}, achieving notable progress. The emergence of agentic frameworks, such as Hilbert~\cite{varambally2025hilbertrecursivelybuildingformal} and Seed-Prover-V1~\cite{chen2025seed}, achieves notable progress by enabling multi-agent theorem decomposition and sub-goal proofs. The latest works apply agentic RL to push LLMs' formal reasoning capability closer to natural language proficiency~\cite{chen2025seed15}. Despite these advancements, formal reasoning in physics remains an underexplored domain, representing a significant opportunity for future research.

\subsection{LLM for Physics Reasoning}\label{related_works:phys}

With the rapid development of general reasoning capabilities in LLMs, researchers are actively exploring the application of these models in more diverse fields~\cite{wang2025let}. Among them, physics reasoning is one key field that receives significant attention. In the context of benchmarks, early comprehensive benchmarks, such as SciBench~\cite{wang2023scibench} and GPQA~\cite{rein2024gpqa}, evaluate college-level scientific problem-solving across multiple scientific fields, including physics. More recently, physics benchmarks have emerged at multiple difficulty levels: UGPhysics~\cite{xu2025ugphysics} presents 5,520 undergraduate-level bilingual problems that advanced thinking models are hard to solve; OlympiadBench~\cite{he2024olympiadbench} introduces 8,476 Olympiad-level problems with multi-module inputs; and recent HiPhO~\cite{yu2025hipho} compiles the latest 13 Physics Olympiad exams in 2024-2025 with human-aligned evaluation.

On the model training side, researchers began exploring the potential for LLMs as physics reasoning tools from an early stage. Early works have demonstrated that LLMs can solve complex word problems that require calculation and inference~\cite{ding2023using}. Such capability can be further enhanced by Reinforcement Learning from Human Feedback (RLHF)~\cite{anand2024enhancing} or simple multi-agent collaboration~\cite{pang2025physics}. Recent works apply RLVR on natural language physics problems, with P1~\cite{chen2025p1} achieving gold-level IPhO performance. However, with a lack of datasets and training methods, developing LLMs for formal physics reasoning is relatively understudied currently~\cite{li2025lean4physics}.

\section{Methodology}
\label{sec:method}


\subsection{Seed Dataset Collection}
\label{sec:method_seed_dataset_collection}
 
    We construct a lemma–proof dataset from the PhysLean GitHub repository~\cite{tooby2025heplean} by extracting all provable lemmas from \texttt{.lean} files along with their preceding formal headers. The lemma statements with context serve as inputs, while the corresponding proof scripts serve as outputs. We filter the samples to retain only those with a total length under 4,096 tokens. The resulting corpus contains over 3,000 examples, which are randomly split into training and test sets at approximately a 9:1 ratio, yielding 2,933 training and 250 test instances. The dataset spans a broad range of domains in physics and mathematics, encompassing classical and modern physics (e.g., classical mechanics, electromagnetism, quantum mechanics, and relativity) as well as advanced theoretical areas such as quantum field theory, string theory, and mathematical foundations.
    An example of the collected data is shown in Figure~\ref{fig:sample-data}.
\subsection{Synthetic Data Generation}
    

    To augment our dataset, we construct a conjecture generation and verification pipeline inspired by STP \citep{dong2025stpselfplayllmtheorem}. Specifically, we treat our initial data as seed data, denoted as $D_\text{seed} = \{(h_i, l_i, p_i)\}_{i=1}^{N}$, where $h_i$, $l_i$, and $p_i$ are the header, lemma, and corresponding proof of the $i^{\text{th}}$ sample, and $N$ is the total number of seed examples. For each sample, we use Claude-4.5-Sonnet \citep{anthropic2025sonnet45} to generate 10 conjectures by providing the header--lemma pairs $(h_i, l_i)$, yielding 29,330 candidate statements. 
    The prompting template is provided in Figure~\ref{fig:claude-prompt} in the Appendix.

After collecting the conjectures, we apply a two-stage pipeline to select well-formed and correct statements. We first examine the \textbf{syntactic correctness} of each conjecture. Specifically, for each conjecture $c_{ij}$, we append it to the corresponding header $h_i$ to form $D_c = \{(h_i, c_{ij}) \mid i = 1, \ldots, N, \; j = 1, \ldots, 10\}$ and use the Lean verifier to check whether the statement is well-formed. This includes verifying that all variables are properly defined and that all referenced definitions and theorems exist. After this step, 6,971 conjectures remain, corresponding to a retention rate of 23.8\%.


The second stage examines the \textbf{provability} of conjectures. Given a conjecture with its corresponding header, we leverage DeepSeek-Prover-V2-7B~\citep{ren2025deepseekproverv2}, Kimina-Prover-Distill-8B~\citep{wang2025kiminaprover}, and Goedel-Prover-V2-8B~\citep{lin2025goedelproverv2}, to generate 16 proofs, producing response samples $\{(h_i, c_{ij}, r_p)\}_{p=1}^{16}$. A conjecture is deemed provable if
$$
\exists \, p, \, 1 \leq p \leq 16 : \texttt{Verify}(h_i, c_{ij}, r_p) = \texttt{True}
$$
where \texttt{Verify} denotes the Lean verification result. This process yields 2,608 verified conjectures, representing an overall pipeline yield rate of 8.9\%, which is comparable to STP \citep{dong2025stpselfplayllmtheorem}. Combining these with the 2,933 seed training examples in Section~\ref{sec:method_seed_dataset_collection} produces a total of 5,541 training instances for our experiments.


Notably, we also compared different proprietary models, including GPT-5 \citep{openai2025gpt5} and Gemini-2.5-Pro \citep{google2025gemini25}. However, the syntactically correct rates of their generated conjectures were substantially lower than those produced by Claude. We additionally explored the approach of generating conjectures in natural language and converting them to Lean4 statements using an auto-formalizer. However, auto-formalizers fail at this task due to the difficulty of identifying a uniform header, which is caused by the complex dependencies in physical statements. Consequently, this approach also yielded low success rates.

\subsection{Self-Evolving Pipeline}\label{sec:self_evolving_pipeline}

We conduct Reinforcement Learning (RL) to lift the performance on physics domain.
Specifically, our experiments are mainly based on Group Relative Policy Optimization (GRPO) \citep{shao2024deepseekmathgrpo}.
For each prompt $x$ in the training set, they sample $G$ (Group size) responses during the rollout stage, and optimize the following objective:


\begin{align*}
\mathcal{J}_{\text{GRPO}}(\theta) &= \mathbb{E}_{x \sim \mathcal{D}, \{y_i\}_{i=1}^G \sim \pi_{\theta_{\text{old}}}(\cdot|x)} \\
&\quad \left[ \frac{1}{G} \sum_{i=1}^{G} \frac{1}{|y_i|} \sum_{t=1}^{|y_i|} \min \left( w_{i,t}(\theta) \hat{A}_{i,t}, \right. \right. \\
&\quad \left. \left. \text{clip}\left(w_{i,t}(\theta), 1-\varepsilon, 1+\varepsilon\right) \hat{A}_{i,t} \right) \right] \\
&\quad - \beta \, \mathbb{D}_{\text{KL}}\left( \pi_\theta \| \pi_{\theta_{\text{ref}}} \right),
\end{align*}

where $y_i$ is the $i^{\text{th}}$ generated sequence of tokens, $\varepsilon$ is the clip ratio. The importance ratio $w_{i,t}(\theta)$ and the advantage $\hat{A}_{i,t}$ are calculated as follows:

\begin{equation*}
w_{i,t}(\theta) = \frac{\pi_\theta(y_{i,t}|x, y_{i,<t})}{\pi_{\theta_{\text{old}}}(y_{i,t}|x, y_{i,<t})},
\end{equation*}

\begin{equation*}
\hat{A}_{i,t} = \hat{A}_i = \frac{r(x, y_i) - \text{mean}\left(\{r(x, y_i)\}_{i=1}^G\right)}{\text{std}\left(\{r(x, y_i)\}_{i=1}^G\right)},
\end{equation*}
respectively. And all the tokens in $y_i$ share the same advantage as $\hat{A}_{i,t}$.

The reward signal $r(x, y_i)$ is provided by the Lean verifier to guide the reinforcement learning process. Specifically, a score of $1$ or $0$ is presented to indicate whether the proof is correct or not. Because of the symbolic nature of Lean, all verified proofs with reward $1$ are correct, with no hallucination at all, which allows the model to learn the foundation of physics in a concrete and rigorous way.
To further reduce the difficulty of the learning process, curriculum learning is employed, where the input statements (conjectures) are sorted based on their ground-truth proof lengths. This enables easy-to-hard learning, encouraging prover models to proceed in a bottom-up manner.


\section{Experiments}
\label{sec:experiments}

\begin{table*}[t]
    \centering
    \small
    \resizebox{\textwidth}{!}{%
    \begin{tabular}{lcccccc}
        \toprule
        \textbf{Method}                     &  \textbf{Classical}   & \textbf{Particle \& String} & \textbf{Relativity} &
        \textbf{Quantum Field Theory}&
        \textbf{Overall} \\
        \midrule
        \textit{Proprietary Models} \\
        \midrule
        \textbf{GPT-5 \citep{openai2025gpt5}}   & 37.3\%  & 13.4\%  & 21.3\%  & 35.2\% & 26.4\% \\
        \textbf{Claude-4.5-Sonnet \citep{anthropic2025sonnet45}} & 52.9\%  & 19.4\%  & 29.5\% & \textbf{39.4\%} & 34.4\% \\
        
        \midrule
        \textit{Formal Math Provers} \\
        \midrule
        \textbf{Kimina-Prover-Distill-8B~\citep{wang2025kiminaprover}}                &  35.3\%                  & 14.9\%             & 29.5\% & 22.5\% & 24.8\% \\
        \textbf{Goedel-Prover-V2-8B~\citep{lin2025goedelproverv2}}                 & 49.0\%                  & 19.4\%              & 34.4\% & 28.2\% & 31.6\% \\
        \textbf{Deepseek-Prover-V2-7B~\citep{ren2025deepseekproverv2}}                        & \multirow{1}{*}{54.9\%}                 & 23.9\%              & 37.7\% & 25.4\% & 34.0\% \\
        \midrule
        \textit{Formal Physics Provers} \\
        \midrule
        \textbf{\ourmodel{}}       & \textbf{58.8\% (+3.9\%)}                  & \textbf{26.9\% (+3.0\%)}              & \textbf{39.3\% (+1.6\%)} & 26.8\% (+1.4\%) & \textbf{36.4\% (+2.4\%)} \\
        \bottomrule
    \end{tabular}}
    \caption{\textbf{Main experimental results}. We evaluate all the models on \datasetname{} test set, which includes Classical, Particle \& String, Relativity, and Quantum Field Theory domains. The pass@16 accuracy is reported.}
    \label{tab:main}
\end{table*}

To evaluate our methodology, we use \datasetname{} to train popular Lean-based formal mathematics provers. Our experiments reveal that strong mathematical reasoning models exhibit notable limitations when handling formal physics problems, underscoring the importance of domain-specific formal datasets and self-evolving strategies.

\subsection{Experimental Setup}

\subsubsection{Dataset and Tasks}



Model performance is evaluated on the test set of \datasetname{}, which shares the same source as the training set with a $9{:}1$ train-test split. To ensure fair comparison across models with different context lengths, we retain only samples with prompt lengths under $4{,}096$ tokens, resulting in $250$ lemmas in the final evaluation set.

For finer-grained analysis, we organize the test samples into four physics categories: Classical \& Foundational Physics, Particle \& String Physics, Relativity \& Spacetime, and Quantum Field Theory. This classification reflects distinct theoretical frameworks and varying levels of required domain expertise. Further details are provided in Appendix~\ref{appendix:physicslean_test}.

\subsubsection{Models and Baselines}


We compare several popular open-source prover models, including DeepSeek-Prover-V2-7B~\citep{ren2025deepseekproverv2}, Kimina-Prover-Distill-8B~\citep{wang2025kiminaprover}, and Goedel-Prover-V2-8B \citep{lin2025goedelproverv2}, all of which are strong formal theorem provers tailored for mathematical domains. Since DeepSeek-Prover-V2-7B performs the best among them, our experiments will focus on training the DeepSeek prover to push the boundaries of open-source models.



For baselines, we first report the performance of DeepSeek-Prover-V2-7B, Kimina-Prover-Distill-8B, and Goedel-Prover-V2-8B without any additional training. We also include comparisons with strong proprietary systems, namely GPT-5 \citep{openai2025gpt5} and Claude-4.5-Sonnet \citep{anthropic2025sonnet45}. For all baselines, we use a fixed sampling budget and report pass@16 accuracy, ensuring fair comparison under a consistent inference budget. For open-source provers, we use the prompt template provided in Appendix~\ref{appendix:prompt_template}. For proprietary models, we employ a tailored Chain-of-Thought (CoT) \citep{wei2023chainofthoughtpromptingelicitsreasoning} prompt to encourage step-by-step reasoning before generating the final proof.

\subsection{Implementation Details}
\label{sec:implement}
We directly apply Reinforcement Learning starting from the DeepSeek-Prover-V2-7B using verl \citep{Sheng_2025}. Specifically, we apply GRPO with rule-based rewards \citep{lambert2025tulu3pushingfrontiers,deepseekai2025deepseekr1incentivizingreasoningcapability} to guide the self-evolving training. In particular, the Lean verifier with version \texttt{4.20.0} is integrated into the verl framework for verifying the proofs. The reward score for each trajectory is calculated as follows:
\begin{equation*}
r(x, y_i) = 
\begin{cases}
1 & \text{if } \text{Verify}(x, y_i) = \text{True} \\
0 & \text{otherwise}
\end{cases}
\end{equation*}
Additionally, if the proof contains \textit{\texttt{sorry, admit, or apply?}} keywords, we directly assign a 0 for the reward score to avoid reward hacking.
Furthermore, to allow a smooth transition of difficulties during the learning process, curriculum learning \citep{parashar2025curriculumreinforcementlearningeasy} is employed by sorting the lemma based on their ground-truth proof lengths.



We train all models on 8$\times$H200 GPUs with a constant learning rate of $1e^{-6}$ and a batch size of 256 for 2 epochs, where the training takes approximately 8 hours.
Notably, we do not have a warm-up stage with Supervised Fine-Tuning (SFT) because it degrades performance. This behavior is investigated and further analyzed in Section~\ref{sec:sft}.
We also investigate the Rejection-Sampling Fine-tuning method \citep{yuan2023scalingrelationshiplearningmathematical,dong2023raftrewardrankedfinetuning} in Section~\ref{sec:sft}.

\begin{figure}[htbp]
\centering
\begin{minipage}[t]{0.48\textwidth}
\begin{tcolorbox}[
  title=Header and Context Lemmas,
  colback=white,
  colframe=black!80,
  coltitle=white,
  fonttitle=\bfseries,
  arc=2mm
]
\begin{lstlisting}
|b|import|b| PhysLean.QFT.PerturbationTheory.WickAlgebra.NormalOrder.Lemmas
|b|import|b| PhysLean.QFT.PerturbationTheory.WickAlgebra.TimeOrder

|b|namespace|b| FieldSpecification
|b|variable|b| {F__ : FieldSpecification}

|b|def timeContract|b| (φ ψ : F__.FieldOp) : F__.WickAlgebra :=
    T__(ofFieldOp φ * ofFieldOp ψ) - T__(ofFieldOp φ * ofFieldOp ψ)

|b|lemma timeContract_eq_superCommute|b| (φ ψ : F__.FieldOp) :
    timeContract φ ψ = if timeOrderRel φ ψ then [anPart φ, ofFieldOp ψ]__s
    else S__(F__ |>__s φ, F__ |>__s ψ) .. [anPart ψ, ofFieldOp φ]__s := by
  split_ifs
  .. rename_i h
    rw [timeContract_of_timeOrderRel _ _ h]
  .. rename_i h
    rw [timeContract_of_not_timeOrderRel_expand _ _ h]

lemma timeContract_zero_of_diff_grade (φ ψ : F__.FieldOp) (h : (F__ |>__s φ) ≠ (F__ |>__s ψ)) :
    timeContract φ ψ = 0 := by
  by_cases h1 : timeOrderRel φ ψ
  .. rw [timeContract_of_timeOrderRel _ _ h1]
    rw [|b|superCommute_anPart_ofFieldOpF_diff_grade_zero|b|]
    exact h
  .. rw [timeContract_of_not_timeOrderRel _ _ h1]
    rw [timeContract_of_timeOrderRel _ _ _]
    rw [superCommute_anPart_ofFieldOpF_diff_grade_zero]
    simp only [instCommGroup.eq_1, smul_zero]
    exact h.symm
    have ht := IsTotal.total (r := F__.timeOrderRel) φ ψ
    simp_all
    ................
\end{lstlisting}
\end{tcolorbox}
\end{minipage}
\hfill
\begin{minipage}[t]{0.48\textwidth}
\begin{tcolorbox}[
  title=DeepseekProver + PhysProver,
  colback=white,
  colframe=black!80,
  coltitle=white,
  fonttitle=\bfseries,
  arc=2mm
]
\begin{lstlisting}
/-- This lemma claims that the normal ordering of the time-contracted product of any two field operators φ and ψ is identically zero.-/
|b|lemma|b| normalOrder_timeContract (φ ψ : F__.FieldOp) :
    N__(timeContract φ ψ) = 0 := by
  rw [|b|timeContract_eq_superCommute|b|]
  split_ifs <;> simp_all [|b|timeContract|b|, |b|superCommute_anPart_ofFieldOpF_diff_grade_zero|b|,
    smul_eq_zero, map_zero]
\end{lstlisting}
\end{tcolorbox}
\end{minipage}
\hfill
\begin{minipage}[t]{0.48\textwidth}
\begin{tcolorbox}[
  title=DeepseekProver,
  colback=white,
  colframe=black!80,
  coltitle=white,
  fonttitle=\bfseries,
  arc=2mm
]
\begin{lstlisting}
|b|lemma|b| normalOrder_timeContract (φ ψ : F__.FieldOp) :
    N__(timeContract φ ψ) = 0 := by
  rw [|b|timeContract_eq_superCommute|b|]
  split_ifs
  .. rw [|r|normalOrder_ofFieldOp_pair_eq_zero|r|]
    <;> simp_all [|r|timeOrderRel_of_isContraction|r|]
  .. rw [normalOrder_anPart_ofFieldOp_smul]
    <;> simp_all [timeOrderRel_of_isContraction]

\end{lstlisting}
\end{tcolorbox}
\end{minipage}
\caption{Successful examples from the \ourmodel{} and failed proofs from the base model for the same statements. \ourmodel{} demonstrates better in-context learning ability to make good usage of lemmas.}
\label{fig:examples}
\end{figure}

\subsection{Experiment Results}


Our experimental results are presented in Table~\ref{tab:main}. We first observe that all existing models achieve relatively low scores despite their proficiency in mathematical theorem proving, with none exceeding 40\% accuracy. Notably, even small open-source theorem prover models exhibit competitive accuracy compared to the latest proprietary systems, such as Claude-4.5-Sonnet and GPT-5. However, proprietary models demonstrate different strengths across physical domains compared to their open-source counterparts. For instance, all open-source provers achieve below 30\% accuracy on Quantum Field Theory, whereas proprietary models exceed 35\%. This suggests that proprietary and open-source models may be trained on different mixtures of physics data.
We also investigated context length in the Quantum Field Theory category and found that the average length is one-third longer than in other domains. These findings align with those of \citet{li2025lean4physics}, suggesting that larger models such as Claude demonstrate superior in-context learning ability, thereby achieving better performance than open-source models.

Our trained model, PhysProver, substantially surpasses its formal mathematics prover counterparts, consistently achieving gains across all categories. Specifically, on the most challenging domains—Particle \& String Physics—where all baselines exhibit low accuracy, our model still yields a notable improvement of 3.0\%. These results demonstrate the effectiveness of extending a mathematics prover to physics domains with only small high-quality datasets. Moreover, the continued performance gains suggest that current provers are far from saturated, indicating that constructing high-quality physics-specific datasets remains a promising direction 

On top of that, the small 7B-sized PhysProver model outperforms both GPT-5 and Claude-4.5-Sonnet in terms of overall performance, which shows the huge potential of small expert models in specific domains of formal physics theorem proving. This provides a promising path toward efficient training of physics prover models.

\begin{table*}[t]
\centering
\begin{tabular}{cc rr}
\toprule
\multicolumn{2}{c}{Problem Category} &  \textbf{Deepseek-Prover-V2} & \textbf{Deepseek-Prover-V2 + \datasetname{}} \\
& & Pass@16 & Pass@16 \\
\midrule
& IMO & 4/20 = 20.0\% & 4/20 = 20.0\% \\
Olympiad & AIME & 8/15 = 53.3\% & 7/15 = 46.7\% \\
& AMC & 25/45 = 55.6\% & 25/45 = 55.6\% \\
\midrule
\multirow{2}{*}{MATH} & Algebra & 63/70 = 90.0\% & 65/70 = 92.9\% \\
& Number Theory & 51/60 = 85.0\% & 53/60 = 88.3\% \\
\midrule
& Algebra & 8/18 = 44.4\% & 8/18 = 44.4\% \\
Custom & Number Theory & 4/8 = 50.0\% & 4/8 = 50.0\% \\
& Induction & 4/8 = 50.0\% & 4/8 = 50.0\% \\
\midrule
\multicolumn{2}{c}{Overall Pass Rate} & 167/244 = 68.4\% & \textbf{170/244 = 69.7\%} \\
\bottomrule
\end{tabular}
\caption{\textbf{Out-of-Distribution Generalization} in Formal Math Proving on MiniF2F-Test \citep{zheng2022minif2fcrosssystembenchmarkformal}.}
\label{tab:minif2f}
\end{table*}

\section{Analysis}
\subsection{Improved In-Context Learning Through Reinforcement Learning}

In this subsection, we provide a detailed analysis of the performance gains achieved by \ourmodel{} through a comparative examination of proofs generated by the baselines and our model. Figure~\ref{fig:examples} presents an illustrative example from our test set along with the corresponding generations. The header and lemmas constitute the context for physical theorem proving, where the lemmas serve as auxiliary tools during the proof process.

We observe that \ourmodel{} consistently makes correct use of functions and lemmas, with successful applications highlighted in \textcolor{blue}{blue}. For instance, to prove the given conjecture, it first applies \textit{timeContract\_eq\_superCommute}, followed by the function \textit{timeContract}. Subsequently, the model correctly invokes \textit{superCommute\_anPart\_ofFieldOpF\_diff\_grade\_zero}, demonstrating effective utilization of contextual information. By synthesizing the knowledge provided in the context, \ourmodel{} successfully completes the proof.

In contrast, while the base model initially applies \textit{timeContract\_eq\_superCommute} correctly, it subsequently generates hallucinated content, including non-existent lemmas such as \textit{normalOrder\_ofFieldOp\_pair\_eq\_zero} and \textit{timeOrderRel\_of\_isContraction} (marked in \textcolor{red}{red}). These observations suggest that the reinforcement learning process on \datasetname{} enhances performance by enabling the model to better leverage contextual information and comprehend domain-specific terminology. This finding also accounts for the low accuracy observed across all base models: their unfamiliarity with physics-specific lemmas and contextual structures impedes their ability to effectively utilize these resources for proof completion.

\subsection{Out-of-Distribution Generalization}

Surprisingly, we also observe that training on physics-centered problems yields notable generalization improvements in formal mathematical theorem proving. In this subsection, we evaluate our trained model on MiniF2F-Test \citep{zheng2022minif2fcrosssystembenchmarkformal}, which comprises 244 Lean4 statements in the mathematics domain, ranging from high school competition problems to elementary undergraduate-level proofs. We partition the dataset into several categories following \citet{ren2025deepseekproverv2}. For each statement in MiniF2F-Test, we prompt both the baseline and our trained model to generate 16 trajectories and compute pass@16 accuracy. We use the same prompt template from the DeepSeek website \footnote{\url{https://huggingface.co/deepseek-ai/DeepSeek-Prover-V2-7B}}.

As shown in Table~\ref{tab:minif2f}, \ourmodel{} overall achieve comparable performance and even surpass their base versions. It is worth noticing that the improvement is not consistent across all categories. For example, our model demonstrates meaningful gains on medium-level problems from MATH \citep{hendrycks2021measuringmathematicalproblemsolving}. Conversely, more challenging Olympiad-level problems may not benefit from GRPO training, as performance drops in the AIME category.
These results reveal both the intrinsic connections and distinctions between mathematical and physical theorem proving in Lean4. In general, training on physics problems can enhance mathematical reasoning capabilities. However, difficult mathematics problems may demand substantially different problem-solving skills that cannot be directly acquired from physics-based training.









\section{Revisiting the Role of Supervised Fine-tuning}
\label{sec:sft}

\begin{table*}[t]
    \centering
    \small
    \resizebox{\textwidth}{!}{%
    \begin{tabular}{lccccccc}
        \toprule
        \textbf{Method}                     & \textbf{Budget}  &  \textbf{Classical}   & \textbf{Particle \& String} & \textbf{Relativity} &
        \textbf{Quantum Field Theory}&
        \textbf{Overall} \\
        \midrule
        \textbf{Deepseek-Prover-V2-7B}                        & \multirow{1}{*}{pass@16} & \textbf{54.9\%}                 & 23.9\%             & 37.7\% & 25.4\% & 34.0\% \\
        \textbf{Deepseek-Prover-V2-7B + Phys SFT}     & pass@16               & 45.1\% \textcolor{red}{(-9.8\%)}                 & 19.4\% \textcolor{red}{(-4.5\%)}   & 26.2\% \textcolor{red}{(-11.5\%)} & 23.9 \textcolor{red}{(-1.5\%)} & 27.6\% \textcolor{red}{(-6.4\%)} \\
        \textbf{Deepseek-Prover-V2-7B + Phys RAFT}     & pass@16               & 52.9\% \textcolor{red}{(-2.0\%)}                 & \textbf{25.4\% \textcolor{blue}{(+1.5\%)}}   & \textbf{41.0\% \textcolor{blue}{(+3.3\%)}} & \textbf{28.2 \textcolor{blue}{(+2.8\%)}} & \textbf{35.6\% \textcolor{blue}{(+1.6\%)}} \\
        \bottomrule
    \end{tabular}}
    \caption{\textbf{Supervised Fine-tuning (SFT) and Rejection-Sampling Fine-tuning (RAFT) on \datasetname{}} of Deepseek-Prover-V2-7B. The pass@16 accuracy drops significantly after SFT, but increases after RAFT.}
    \label{tab:sft}
\end{table*}

\begin{table}[h]
\centering
\begin{tabular}{lcc}
\toprule
 & Training Set  & Test Set \\
\midrule
DS-Prover-SFT & 1.817 & 1.711 \\
DS-Prover-RAFT & 1.321 & \textbf{1.186} \\
DS-Prover-GRPO & \textbf{1.141} & 1.209 \\
\bottomrule
\end{tabular}
\caption{Perplexity on the training set and the test set for DS-Prover-SFT, DS-Prover-RAFT, and DS-Prover-GRPO.}
\label{tab:perplexity}
\end{table}

We additionally investigated whether conducting Supervised Fine-tuning (SFT) on \datasetname{} could enhance model performance on Physics, following standard practice in training specialized LLMs. However, we did not observe any improvement on our test set after SFT. Instead, we observed consistent performance degradation.

Specifically, we first fine-tuned Deepseek-Prover-V2-7B on \datasetname{}, where ground-truth answers were either extracted from the PhysLean library \textbf{written by humans}, or generated by open-source provers with subsequent verifications. The training sample template follows the RL prompt template in~\ref{appendix:prompt_template}, with loss computation restricted to the completion portion. 
We then consider Rejection Sampling Fine-tuning, or Reward-Ranked Fine-tuning (RAFT) \citep{dong2023raftrewardrankedfinetuning,yuan2023scalingrelationshiplearningmathematical}, where we sample Deepseek-Prover-V2-7B on the training set and retain only the correct proofs as our new training set.
We fine-tuned Deepseek-Prover-V2-7B on both training sets for one epoch, using a learning rate of $5e^{-7}$ and a batch size of 32.
They are denoted as DS-Prover-SFT and DS-Prover-RAFT, respectively.

As shown in Table~\ref{tab:sft}, for the DS-Prover-SFT, we observe consistent performance degradation across all categories, with an average accuracy decline of 6.4\%.
On the contrary, DS-Prover-RAFT demonstrates an overall 1.6\% improvement, with all other 3 categories increasing except Classical Physics.
We attribute this performance difference to the distributional properties of the training data. The original \datasetname{} consists of human-written examples, which are Out-of-Distribution (OOD) with respect to the model's generation capabilities. In contrast, Rejection Sampling yields In-Distribution (ID) data that more closely aligns with the model's output distribution. Consequently, ID data may be easier for the model to learn from, leading to improved performance.

To gain a closer look into this phenomenon, probing experiments are conducted to compare the uncertainty of the SFT model, RAFT model (Table~\ref{tab:sft}), and the GRPO model from our main experiments.

To assess model uncertainty on both training and test data, we measured the average perplexity of sampled responses conditioned on input prompts. Given a prompt $x$ from either the training or test set, we sampled $K=16$ responses $y_k$ from the model and computed the mean perplexity across these samples. We randomly selected 50 samples from each of the training and test sets. The computation is defined as:
\begin{align*}
\overline{PPL}(x) = \frac{1}{K} \sum_{k=1}^{K} PPL(y^{(k)}), \quad y^{(k)} \sim p_\theta(\cdot \mid x)
\end{align*}
where
\begin{align*}
PPL(y) = \exp\left(-\frac{1}{|y|}\sum_{t=1}^{|y|} \log p_\theta(y_t \mid y_{<t}, x)\right).
\end{align*}
This metric captures the model's self-uncertainty: lower values indicate that the model generates responses it considers likely and more relevant to the input, while higher values suggest greater variability or unfamiliarity with the prompt.

As shown in Table~\ref{tab:perplexity}, the findings demonstrate that the average perplexity for DS-Prover-GRPO and DS-Prover-RAFT is substantially lower than that of DS-Prover-SFT on both the training and test sets, which explains why GRPO and RAFT improve the performance while SFT does not.
These results suggest that although supervised fine-tuning directly maximizes the probability of target tokens, it does not necessarily reduce model uncertainty, particularly for models such as DeepSeek-Prover that have already undergone extensive domain-specific (MATH) training. This observation offers an important insight for further improving expert models: supervised fine-tuning may not always be necessary or optimal. 
On the contrary, using the Rejection Sampling Fine-tuning method to collect and fine-tune on In-Distribution (ID) data could be a pratical solution.
Additionally, direct application of reinforcement learning can serve as a viable alternative, particularly in low-resource settings.
We also explore Reinforcement Learning after Rejection-Sampling Fine-tuning in Appendix~\ref{appendix:rft+rl}, but do not observe improvements.

\section{Conclusion}
\label{sec:conclusion}

In this paper, we present the first systematic effort to advance formal theorem proving in the physical domain. We first introduce \datasetname{}, a dataset of physical theorems formalized in Lean4, along with a conjecture formulation pipeline for generating valid and correct conjectures. By applying Reinforcement Learning with Verifiable Rewards (RLVR) to an open-source state-of-the-art theorem prover, our \ourmodel{} achieves consistent \textbf{2.4\%} improvements across physical sub-domains such as Quantum Field Theory using only 5K samples. The model also demonstrates over \textbf{1\%} improvement on the out-of-distribution MiniF2F-test benchmark, highlighting strong generalization capability.
Our work bridges a critical gap between formal theorem proving in mathematics and its application to the physical sciences. We will publicly release our dataset and models to facilitate future research in this direction.
\section{Limitations}

Our work has several limitations that we acknowledge and hope to address in future research.
First, due to computational resource constraints, we were unable to collect more data or scale the conjecture generation process to a larger extent. As noted in Section 3.2, our synthetic data pipeline has a yield rate of only 8.9\%, meaning that a substantial portion of generated conjectures are filtered out during validity and correctness verification. Scaling up the generation process would require significantly more compute for both the LLM-based conjecture generation and the multi-prover verification stage, which was beyond our current budget.
Additionally, our dataset is derived solely from the PhysLean repository, which, while comprehensive, may not cover all areas of physics uniformly. Certain specialized domains may be underrepresented, potentially limiting the model's applicability to the full breadth of physical theorem proving.



\bibliography{custom}

\appendix
\section{Author Contributions}

This work stems from all authors’ valuable contributions and close collaborations.

\noindent \textbf{HZ}, together with RW, develops the Lean verification pipeline. HZ constructs the conjecture generation pipeline to generate conjectures, develops the RL training pipeline integrated with Lean evaluation, conducts all experiments presented in this paper, and, based on the skeleton provided by RP, writes most of the paper except for the related work section and performs polishing.

\noindent \textbf{RW}, provides the initial main idea for the paper. RW builds most of the Lean verification pipeline and collaborates closely with HZ on both data generation and model training. Additionally, RW also contributed to analysis of case study. In terms of paper writing, RW contributed to the Introduction and Related Work sections and modified the other sections in terms of logical and conceptual coherence.

\noindent \textbf{RP}, together with RW, initiates the project, drafting proposals to ensure computational resources, debugging the Lean evaluation pipeline, and driving collaboration efforts between different members. RP also provides the skeleton of the paper, the first draft of the experimental section, draws Figure~\ref{fig:main}, and polishes the paper by supplementing Section~\ref{sec:self_evolving_pipeline} and improving coherence/grammatical correctness.

\noindent \textbf{WW}, together with RW, provides the initial Lean dataset, consisting of provable theorems and their proof scripts extracted from the \textsc{PhysLean} library. WW also conducts preliminary explorations on SFT experimental design and the verification pipeline. In terms of paper writing, the collected dataset is described in Section~3.1 Seed Dataset Collection. 

\noindent \textbf{BM}, together with HZ and RW, reviewed case studies for both open-source and proprietary model responses. BM contributed physics expertise, ensuring the mathematics accuracy and physics soundness of all model outputs.

\noindent \textbf{TZ} supports the work and provides computational resources, guidance, and suggestions for experiment design and paper writing.

\section{PhysProver Categories}
\label{appendix:physicslean_test}

\begin{table*}[t]
    \centering
    \small
    \resizebox{\textwidth}{!}{%
    \begin{tabular}{lccccccc}
        \toprule
        \textbf{Method}                     & \textbf{Budget}  &  \textbf{Classical}   & \textbf{Particle \& String} & \textbf{Relativity} &
        \textbf{Quantum Field Theory}&
        \textbf{Overall} \\
        \midrule
        \textbf{DeepSeek-Prover-V2-7B}                        & \multirow{1}{*}{pass@16} & 54.9\%                & 23.9\%             & 37.7\% & 25.4\% & 34.0\% \\
        \textbf{+RAFT}     & pass@16               & 52.9\% \textcolor{red}{(-2.0\%)}                 & 25.4\% \textcolor{blue}{(+1.5\%)} & \textbf{41.0\% \textcolor{blue}{(+3.3\%)}} & \textbf{28.2 \textcolor{blue}{(+2.8\%)}} & 35.6\% \textcolor{blue}{(+1.6\%)} \\
        \textbf{RAFT+GRPO Step 5}                        & \multirow{1}{*}{pass@16} & 52.9\%   \textcolor{red}{(-2.0\%)}              & 23.9\%  \textcolor{blue}{(+0.0\%)}          & \textbf{41.0\% \textcolor{blue}{(+3.3\%)}} & \textbf{28.2 \textcolor{blue}{(+2.8\%)}} & 35.2\% \textcolor{blue}{(+1.2\%)}\\
        \textbf{RAFT+GRPO Step 10}                        & \multirow{1}{*}{pass@16} & 52.9\% \textcolor{red}{(-2.0\%)}                & 25.4\% \textcolor{blue}{(+1.5\%)}            & 39.3\% \textcolor{blue}{(+1.6\%)} & 26.8\% \textcolor{blue}{(+1.4\%)} & 34.8\% \textcolor{blue}{(+0.8\%)} \\
        \textbf{RAFT+GRPO Step 15}                        & \multirow{1}{*}{pass@16} & \textbf{56.9\%}   \textcolor{blue}{(+2.0\%)}              & 25.4\%  \textcolor{blue}{(+1.5\%)}           & 39.3\% \textcolor{blue}{(+1.6\%)} & 26.8\% \textcolor{blue}{(+1.4\%)} & 35.6\% \textcolor{blue}{(+1.6\%)} \\
        \textbf{RAFT+GRPO Step 20}                        & \multirow{1}{*}{pass@16} & 54.9\%  \textcolor{blue}{(+0.0\%)}               & 25.4\%  \textcolor{blue}{(+1.5\%)}           & \textbf{41.0\% \textcolor{blue}{(+3.3\%)}} & \textbf{28.2 \textcolor{blue}{(+2.8\%)}} & \textbf{36.0\% \textcolor{blue}{(+2.0\%)}} \\
        \textbf{RAFT+GRPO Step 25}                        & \multirow{1}{*}{pass@16} & 54.9\%   \textcolor{blue}{(+0.0\%)}               & 25.4\%         \textcolor{blue}{(+1.5\%)}    & 39.3\% \textcolor{blue}{(+1.6\%)} & \textbf{28.2 \textcolor{blue}{(+2.8\%)}} & 35.6\% \textcolor{blue}{(+1.6\%)} \\
        \textbf{RAFT+GRPO Step 30}                        & \multirow{1}{*}{pass@16} & 54.9\%  \textcolor{blue}{(+0.0\%)}                 & \textbf{26.9\%  \textcolor{blue}{(+3.0\%)}}          & 37.7\% \textcolor{blue}{(+0.0\%)} & \textbf{28.2 \textcolor{blue}{(+2.8\%)}} & 35.6\% \textcolor{blue}{(+1.6\%)} \\
        \bottomrule
    \end{tabular}}
    \caption{\textbf{Reinforcement Learning after Rejection-Sampling Fine-tuning (RAFT) on \datasetname{}} of Deepseek-Prover-V2-7B. The pass@16 accuracy does not improve after RAFT.}
    \label{tab:raft+rl}
\end{table*}

The Classical \& Foundational Physics category groups core undergraduate-level subjects, including mathematical methods, classical mechanics, quantum mechanics, statistical mechanics, and electromagnetism. These areas represent foundational discoveries in physics and are primarily textbook-driven, with standardized problem formulations and solution methods.

Particle and String Physics is grouped separately to capture topics centered on high-energy physics and fundamental interactions, often motivated by experimental programs such as those at the Large Hadron Collider. String theory topics are included in this category due to their close conceptual alignment with high-energy theoretical frameworks.

Quantum Field Theory and Relativity are treated as distinct categories due to their advanced mathematical structure and conceptual complexity. Both subjects are typically introduced at the graduate level, with quantum field theory extending quantum mechanics and relativity providing a foundational framework for spacetime and gravitation.

\section{Reinforcement Learning after Rejection-Sampling Fine-tuning}
\label{appendix:rft+rl}



In this section, we investigate the effectiveness of Reinforcement Learning (RL) following Rejection-Sampling Fine-tuning (RAFT). Starting from the RAFT checkpoint described in Section~\ref{sec:sft}, we apply RL using the same hyperparameters and settings as in our main experiments (Section~\ref{sec:implement}). We save checkpoints every 5 steps and report RL performance in Table~\ref{tab:raft+rl}.

We observe that overall performance does not improve beyond 30 steps of GRPO training, with overall accuracy remaining unchanged (both 35.6\%). While the Particle \& String category shows improvement relative to both the base model and the RAFT model, we observe a corresponding performance drop in the Relativity category, indicating inconsistent gains across categories. Throughout the RL process, both overall and category-level accuracies fluctuate without demonstrating a clear trend. Although we observe the best overall results at Step 20, we attribute this to random noise given the training process.
These results suggest that RL after RAFT may not be consistently effective when both stages use the same prompt set. This observation aligns with common practice in LLM training, where different prompt sets are typically employed for fine-tuning and reinforcement learning.

\section{Experimental Details}
\label{appendix:exp_details}

\subsection{Prompt Template for Main Experiments}
\label{appendix:prompt_template}

We list the prompt template for DeepSeek-Prover-V2-7B in Figure~\ref{fig:deepseek-prompt} with a concrete example from \datasetname{}. 
Specifically, in the user round, we append the statement with \texttt{sorry} to act as a task for the model. And in the assistant round, we give the model the entire context of the header and the statement.
We chose this template because we found it stable for the models to generate proof completions, as it allows the model to directly generate the proof part.

The prompt template for Kimina-Prover and Goedel-Prover is exactly the same except for the special tokens which are shown in Figure~\ref{fig:kimina-prompt}.

The prompt template for the proprietary models is different due to the inaccessibility of the actual models.
Instead, we apply the Chain-of-Thought prompting \citep{wei2023chainofthoughtpromptingelicitsreasoning}, and the template is shown in Figure~\ref{fig:proprietary-prompt}

We also list the prompt template for Claude-4.5-Sonnet to generate conjectures in Figure~\ref{fig:claude-prompt}.

\subsection{Conjecture Generation}

We also list an example of the conjecture, given a header and a lemma, which is shown in Figure~\ref{fig:conjecture-example}.
In this example, the conjecture is a variant of the original lemma - Both lemmas are basic algebraic properties of matrix-vector multiplication. 
In other examples we examined, the conjectures consistently fall into predictable categories relative to their preceding lemmas.
These variations typically manifest as: (1) special cases obtained by substituting specific values such as zero, (2) definitional unwrapping where a concept is restated in more explicit form, or (3) algebraic variants that apply the same underlying principle to a different argument.
Since PhysLean formalizes abstract concepts in mathematical physics, many lemmas establish foundational API properties—such as linearity, group action behavior, or interaction with zero elements—and the conjectures naturally complete this API by covering symmetric or boundary cases. Consequently, these conjectures primarily serve to help models become more familiar with the dataset's conventions, naming patterns, and proof structures to improve the accuracy, rather than to evaluate deep mathematical reasoning capabilities.

\begin{figure*}[t]
\caption{An example of the \datasetname{}. The black lines denote the header, the blue lines denote the lemma statement, and the red lines denote the proof.}
\label{fig:sample-data}
\begin{minipage}[t]{0.98\textwidth}
\begin{tcolorbox}[
  title={Example of Header, Statement, and the Proof from PhysLean},
  colback=white,
  colframe=black!80,
  coltitle=white,
  fonttitle=\bfseries,
  arc=2mm
]
\begin{lstlisting}[basicstyle=\ttfamily\small, columns=fullflexible, breaklines=true]

import PhysLean.Mathematics.Fin.Involutions
import PhysLean.QFT.PerturbationTheory.WickContraction.ExtractEquiv
import PhysLean.QFT.PerturbationTheory.WickContraction.Involutions

open FieldSpecification

variable {F__ : FieldSpecification}

namespace WickContraction

variable {n : N__} (c : WickContraction n)

open PhysLean.List
open FieldStatistic
open Nat

def IsFull : Prop := c.uncontracted = empty__

instance : Decidable (IsFull c) := decEq c.uncontracted empty__

/-- This theorem states that the configuration c is full if and only if, for every index i in the finite set of size n, the involution function associated with c does not map i to itself.-/

|b|lemma isFull_iff_equivInvolution_no_fixed_point :
IsFull c <-> ∀ (i : Fin n), (equivInvolution c).1 i ≠ i := by |b|
|r|simp only [IsFull, ne_eq]
rw [Finset.eq_empty_iff_forall_notMem]
simp [equivInvolution, toInvolution, uncontracted] |r|

\end{lstlisting}
\end{tcolorbox}
\end{minipage}

\end{figure*}

\begin{figure*}[t]
\caption{Prompt template for DeepSeek Prover with a concrete example}
\label{fig:deepseek-prompt}
\begin{minipage}[t]{0.98\textwidth}
\begin{tcolorbox}[
  title=DeepSeek Prover Prompt Template,
  colback=white,
  colframe=black!80,
  coltitle=white,
  fonttitle=\bfseries,
  arc=2mm
]
\begin{lstlisting}[basicstyle=\ttfamily\small, columns=fullflexible, breaklines=true]
|b|<|begin_of_sentence|> |b|You are an expert in mathematics, physics and Lean 4.
|b|<|User|>|b|Think about and solve the following problem step by step in Lean 4. -> Directly give the formal statement in Lean 4.

# Problem: This theorem states that the matrix product of the Minkowski metric with itself yields the identity matrix.
# Formal statement:
```lean4
import Mathlib.Algebra.Lie.Classical
import Mathlib.Analysis.Normed.Ring.Lemmas

open Matrix

def minkowskiMatrix {d : NN} : Matrix (Fin 1 ++ Fin d) (Fin 1 ++ Fin d) RR :=
  LieAlgebra.Orthogonal.indefiniteDiagonal (Fin 1) (Fin d) RR

namespace minkowskiMatrix

variable {d : NN}

scoped[minkowskiMatrix] notation "eta" => minkowskiMatrix

/-- This theorem states that the matrix product of the Minkowski metric with itself yields the identity matrix.-/
@[simp]
lemma sq : @minkowskiMatrix d * minkowskiMatrix = 1 :=
  sorry
```

|b|<|Assistant|>|b|
```lean4
import Mathlib.Algebra.Lie.Classical
import Mathlib.Analysis.Normed.Ring.Lemmas

open Matrix

def minkowskiMatrix {d : NN} : Matrix (Fin 1 ++ Fin d) (Fin 1 ++ Fin d) RR :=
  LieAlgebra.Orthogonal.indefiniteDiagonal (Fin 1) (Fin d) RR

namespace minkowskiMatrix

variable {d : NN}

scoped[minkowskiMatrix] notation "eta" => minkowskiMatrix

/-- This theorem states that the matrix product of the Minkowski metric with itself yields the identity matrix.-/
@[simp]
|b|lemma sq : @minkowskiMatrix d * minkowskiMatrix = 1 :=|b|
\end{lstlisting}
\end{tcolorbox}
\end{minipage}

\end{figure*}

\begin{figure*}[t]
\caption{Prompt template for Kimina Prover and Goedel Prover with a concrete example. They are the same as the Deepseek Prover except for the special tokens, such as the BOS token.}
\label{fig:kimina-prompt}
\begin{minipage}[t]{0.98\textwidth}
\begin{tcolorbox}[
  title=Kimina and Goedel Provers Prompt Template,
  colback=white,
  colframe=black!80,
  coltitle=white,
  fonttitle=\bfseries,
  arc=2mm
]
\begin{lstlisting}[basicstyle=\ttfamily\small, columns=fullflexible, breaklines=true]

|b|<|im_start|>system |b|
You are an expert in mathematics, physics and Lean 4.<|im_end|>
|b|<|im_start|>user|b|
Think about and solve the following problem step by step in Lean 4. -> Directly give the formal statement in Lean 4.

# Problem: This theorem states that the matrix product of the Minkowski metric with itself yields the identity matrix.
# Formal statement:
```lean4
import Mathlib.Algebra.Lie.Classical
import Mathlib.Analysis.Normed.Ring.Lemmas

open Matrix

def minkowskiMatrix {d : NN} : Matrix (Fin 1 ++ Fin d) (Fin 1 ++ Fin d) RR :=
  LieAlgebra.Orthogonal.indefiniteDiagonal (Fin 1) (Fin d) RR

namespace minkowskiMatrix

variable {d : NN}

scoped[minkowskiMatrix] notation "eta" => minkowskiMatrix

/-- This theorem states that the matrix product of the Minkowski metric with itself yields the identity matrix.-/
@[simp]
lemma sq : @minkowskiMatrix d * minkowskiMatrix = 1 :=
  sorry
```<|im_end|>
|b|<|im_start|>assistant|b|
```lean4
import Mathlib.Algebra.Lie.Classical
import Mathlib.Analysis.Normed.Ring.Lemmas

open Matrix

def minkowskiMatrix {d : NN} : Matrix (Fin 1 ++ Fin d) (Fin 1 ++ Fin d) RR :=
  LieAlgebra.Orthogonal.indefiniteDiagonal (Fin 1) (Fin d) RR

namespace minkowskiMatrix

variable {d : NN}

scoped[minkowskiMatrix] notation "eta" => minkowskiMatrix

/-- This theorem states that the matrix product of the Minkowski metric with itself yields the identity matrix.-/
@[simp]
|b|lemma sq : @minkowskiMatrix d * minkowskiMatrix = 1 :=|b|

\end{lstlisting}
\end{tcolorbox}
\end{minipage}

\end{figure*}

\begin{figure*}[t]
\caption{Prompt template for proprietary models for the main experiments.}
\label{fig:proprietary-prompt}
\begin{minipage}[t]{0.98\textwidth}
\begin{tcolorbox}[
  title=Proprietary Models Prompt Template,
  colback=white,
  colframe=black!80,
  coltitle=white,
  fonttitle=\bfseries,
  arc=2mm
]
\begin{lstlisting}[basicstyle=\ttfamily\small, columns=fullflexible, breaklines=true]

Complete the following Lean 4 code:

```lean4
{|b|THE LEAN4 HEADERS AND STATEMENT|b|}
```

Before producing the Lean 4 code to formally prove the given theorem, provide a proof plan outlining the main proof steps and strategies.

The plan should highlight key ideas, intermediate lemmas, and proof structures that will guide the construction of the final formal proof.

\end{lstlisting}
\end{tcolorbox}
\end{minipage}

\end{figure*}



\begin{figure*}[t]
\caption{Prompt template for Claude-4.5-Sonnet to generate formal conjectures}
\label{fig:claude-prompt}
\begin{minipage}[t]{0.98\textwidth}
\begin{tcolorbox}[
  title=Claude-4.5-Sonnet Prompt Template for Conjecture Generation,
  colback=white,
  colframe=black!80,
  coltitle=white,
  fonttitle=\bfseries,
  arc=2mm
]
\begin{lstlisting}[basicstyle=\ttfamily\small, columns=fullflexible, breaklines=true]

You are an expert in mathematics, physics and Lean 4.
You are provided a context, a lemma, and a proof. Your task is to generate a list of 10 related physics conjecture in formal language based on the context and the seed language statements.

The conjectures should be:
1. A meaningful variant of the original theorem: modify hypotheses, generalize structures, or extend scope while keeping the core mathematical insight.
2. Must differ significantly in mathematical content (changed assumptions, stronger/weaker conclusions, or different algebraic structures) but remain recognizably related.
3. The new conjecture should be in formal language.
4. Do not include the proof.

When generating the conjectures, preserve all specific Lean identifiers exactly as they appear in the formal statement. You can also refer to the original formal statement.

Context:
{context}

Natural Language Statement:
{nq}

Original Formal Statement:
{theorem}

Return the final conjectures in JSON format as a dictionary where:
- The key is "conjectures"
- The value is a list of dictionaries
- Each dictionary in the list has a key "statement" whose value is a string containing one conjecture

Please read, understand, and then generate a list of conjectures.

\end{lstlisting}
\end{tcolorbox}
\end{minipage}

\end{figure*}

\begin{figure*}[t]
\caption{An example of the conjecture based on the header and the lemma}
\label{fig:conjecture-example}
\begin{minipage}[t]{0.98\textwidth}
\begin{tcolorbox}[
  title=The Header of the Original Lemma,
  colback=white,
  colframe=black!80,
  coltitle=white,
  fonttitle=\bfseries,
  arc=2mm
]
\begin{lstlisting}[basicstyle=\ttfamily\small, columns=fullflexible, breaklines=true]
import PhysLean.Relativity.PauliMatrices.SelfAdjoint
import Mathlib.RepresentationTheory.Basic
import PhysLean.Relativity.Lorentz.Group.Basic
import Mathlib.Analysis.InnerProductSpace.PiL2

namespace Lorentz

noncomputable section
open Matrix
open MatrixGroups
open Complex

structure ContrMod (d : NN) where
  val : Fin 1 ++ Fin d → ℝ

namespace ContrMod

variable {d : NN}

def toFin1dℝFun : ContrMod d ~= (Fin 1 ++ Fin d → ℝ) where
  toFun v := v.val
  invFun f := leftarrfrightarr
  left_inv _ := rfl
  right_inv _ := rfl

instance : AddCommMonoid (ContrMod d) := Equiv.addCommMonoid toFin1dℝFun

instance : AddCommGroup (ContrMod d) := Equiv.addCommGroup toFin1dℝFun

instance : Module ℝ (ContrMod d) := Equiv.module ℝ toFin1dℝFun

def toFin1dℝEquiv : ContrMod d ~=l[ℝ] (Fin 1 ++ Fin d → ℝ) :=
  Equiv.linearEquiv ℝ toFin1dℝFun

abbrev toFin1dℝ (ψ : ContrMod d) := toFin1dℝEquiv ψ

@[simps!]
def stdBasis : Basis (Fin 1 ++ Fin d) ℝ (ContrMod d) := Basis.ofEquivFun toFin1dℝEquiv

abbrev mulVec (M : Matrix (Fin 1 ++ Fin d) (Fin 1 ++ Fin d) ℝ) (v : ContrMod d) :
    ContrMod d := Matrix.toLinAlgEquiv stdBasis M v

scoped[Lorentz] infixr:73 " *v " => ContrMod.mulVec

\end{lstlisting}
\end{tcolorbox}
\end{minipage}

\begin{minipage}[t]{0.98\textwidth}
\begin{tcolorbox}[
  title=The Original Lemma,
  colback=white,
  colframe=black!80,
  coltitle=white,
  fonttitle=\bfseries,
  arc=2mm
]
\begin{lstlisting}[basicstyle=\ttfamily\small, columns=fullflexible, breaklines=true]
/-- This theorem states that for any two real matrices M and N of dimension (1+d) xx (1+d) and any vector v in ContrMod d, the matrix-vector product of their difference with v equals the difference of their individual matrix-vector products with v.-/
|b|lemma sub_mulVec (M N : Matrix (Fin 1 ++ Fin d) (Fin 1 ++ Fin d) ℝ) (v : ContrMod d) :
    (M - N) *v v = M *v v - N *v v := sorry|b|

\end{lstlisting}
\end{tcolorbox}
\end{minipage}

\begin{minipage}[t]{0.98\textwidth}
\begin{tcolorbox}[
  title=The New Conjecture,
  colback=white,
  colframe=black!80,
  coltitle=white,
  fonttitle=\bfseries,
  arc=2mm
]
\begin{lstlisting}[basicstyle=\ttfamily\small, columns=fullflexible, breaklines=true]

|b|lemma mulVec_zero (M : Matrix (Fin 1 ++ Fin d) (Fin 1 ++ Fin d) ℝ) : M *v (0 : ContrMod d) = 0 := sorry|b|

\end{lstlisting}
\end{tcolorbox}
\end{minipage}

\end{figure*}






\end{document}